\documentclass[conference]{IEEEtran}
\IEEEoverridecommandlockouts
\usepackage{cite}
\usepackage{amsmath,amssymb,amsfonts}
\usepackage{algorithmic}
\usepackage{graphicx}
\usepackage{textcomp}
\usepackage{xcolor}
\usepackage{times}  
\usepackage[hyphens]{url}  
\usepackage{graphicx} 
\usepackage{caption} 

\usepackage{graphicx}
\usepackage{booktabs}
\usepackage{subfigure}
\usepackage{algorithm}
\usepackage{algorithmic}
\usepackage{color}

\usepackage{microtype}
\usepackage{amsmath, amssymb, amscd, amsfonts}
\usepackage{soul}
\usepackage{url}
\usepackage[hidelinks]{hyperref}
\usepackage{amsmath}
\usepackage{algorithm}
\usepackage{algorithmic}
\usepackage{amssymb}
\usepackage{amsfonts} 
\usepackage{multirow}
\usepackage{boldline}
\usepackage{hhline}
\usepackage{xcolor}

\usepackage{makecell}
\usepackage{mathrsfs}
\usepackage[normalem]{ulem}
\useunder{\uline}{\ul}{}
\usepackage[inline]{enumitem}
\usepackage{enumitem}
\usepackage{bbding}
\usepackage{pifont}

\def\BibTeX{{\rm B\kern-.05em{\sc i\kern-.025em b}\kern-.08em
    T\kern-.1667em\lower.7ex\hbox{E}\kern-.125emX}}
\begin{document}

\title{A Knowledge-Enhanced Adversarial Model for Cross-lingual Structured Sentiment Analysis}

\author{
\IEEEauthorblockN{1\textsuperscript{st} Qi Zhang}
\IEEEauthorblockA{\textit{School of Computer Science and Technology} \\
\textit{East China Normal University} \\
Shanghai, China \\
qzhang@stu.ecnu.edu.cn
}
\and
\IEEEauthorblockN{2\textsuperscript{nd} Jie Zhou*\thanks{\quad Jie Zhou is the corresponding author of this paper.}}
\IEEEauthorblockA{\textit{School of Computer Science} \\
\textit{Fudan Univerisity}\\
Shanghai, China \\
jie\_zhou@ica.stc.sh.cn}
\and
\IEEEauthorblockN{3\textsuperscript{rd} Qin Chen}
\IEEEauthorblockA{\textit{School of Computer Science and Technology} \\
\textit{East China Normal University}\\
Shanghai, China \\
qchen@cs.ecnu.edu.cn}
\and
\IEEEauthorblockN{4\textsuperscript{th} Qingchun Bai}
\IEEEauthorblockA{\textit{Shanghai Open University} \\
Shanghai, China \\
qc\_bai@foxmail.com}
\and
\IEEEauthorblockN{5\textsuperscript{th} Jun Xiao}
\IEEEauthorblockA{\textit{Shanghai Open University} \\
Shanghai, China \\
xiaoj@sou.edu.cn}
\and
\IEEEauthorblockN{6\textsuperscript{th} Liang He}
\IEEEauthorblockA{\textit{School of Computer Science and Technology} \\
\textit{East China Normal University}\\
Shanghai, China \\
lhe@cs.ecnu.edu.cn}
}

\maketitle

\begin{abstract}
Structured sentiment analysis, which aims to extract the complex semantic structures such as holders, expressions, targets, and polarities, has obtained widespread attention from both industry and academia. 
Unfortunately, the existing structured sentiment analysis datasets refer to a few languages and are relatively small, limiting neural network models' performance.
In this paper, we focus on the cross-lingual structured sentiment analysis task, which aims to transfer the knowledge from the source language to the target one. 
Notably, we propose a Knowledge-Enhanced Adversarial Model (\texttt{KEAM}) with both implicit distributed and explicit structural knowledge to enhance the cross-lingual transfer. 
First, we design an adversarial embedding adapter for learning an informative and robust representation by capturing implicit semantic information from diverse multi-lingual embeddings adaptively. 
Then, we propose a syntax GCN encoder to transfer the explicit semantic information (e.g., universal dependency tree) among multiple languages. 
We conduct experiments on five datasets and compare \texttt{KEAM} with both the supervised and unsupervised methods.
The extensive experimental results show that our \texttt{KEAM} model outperforms all the unsupervised baselines in various metrics.  
\end{abstract}


\section{Introduction}
\begin{figure}[!t]
\vspace{-2mm}
    \centering
    \includegraphics[scale=0.36]{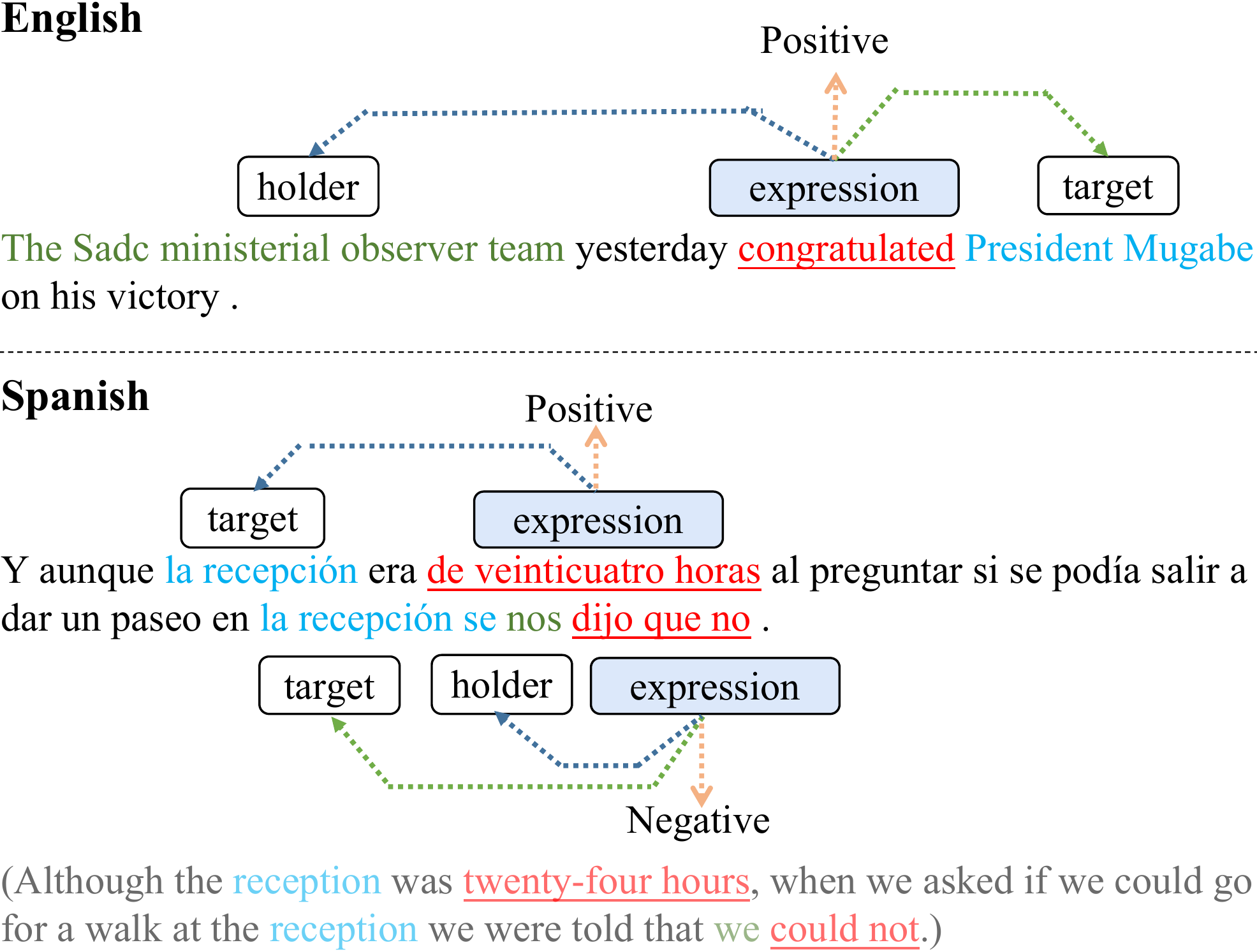}
    \vspace{-1mm}
    \caption{An example of structured sentiment analysis. We translate the Spanish into English in the brackets for ease of understanding.}
    \label{fig:example}
    \vspace{-4mm}
\end{figure}

Sentiment analysis is a fundamental task of natural language processing (NLP), which aims to determine the emotion, sentiment, and opinions about a product, attribute, target, or policy \cite{bakshi2016opinion,zhou2020sentix}. 
Recently, more and more researchers focus on fine-grained sentiment analysis task, such as aspect-based sentiment analysis \cite{pontiki2016semeval,zhou2020sk}, targeted sentiment analysis \cite{mitchell2013open}, and structured sentiment analysis \cite{DBLP:conf/acl/BarnesKOOV20}. 
In this paper, we focus on the task of structured sentiment analysis, which consists of four sub-tasks: 1) expression extraction, 2) target extraction, 3) holder extraction, and 4) sentiment prediction. As shown in Fig. \ref{fig:example}, ``The Sadc ministerial observer team" (holder $h$) expresses a positive sentiment (polarity $p$) towards ``President Mugabe" (target $t$) with ``congratulated" (expression $e$).

Unfortunately, the existing publicly available datasets for structured sentiment analysis refer to a few languages \cite{DBLP:conf/acl/BarnesKOOV20} and are relatively small, which limit the effectiveness of neural network models. 
The annotations for structured sentiment analysis are time-consuming and laborious because they are structured and require a rich label space.
Furthermore, it is challenging to extract complex semantic structures such as expressions, holders, targets, and sentiments, especially for corpus written in under-annotated and under-resourced languages.
To address this problem, we focus on a cross-lingual structured sentiment analysis task, which aims to translate the structural sentiment knowledge from the source language (e.g., English) to the target language (e.g., Spanish).

Previous works about structured sentiment analysis focus on modeling the structure and sentiment representation in mono-lingual \cite{DBLP:conf/acl/BarnesKOOV20,mitchell2013open}.
As a specifical case of fine-grained sentiment analysis task \cite{pontiki2016semeval}, neural-based pipeline and joint algorithms are proposed for this tasks \cite{wang-etal-2016-recursive,he-etal-2019-interactive}. 
They extracted the targets, expressions, and sentiments by converting these tasks into a sequence labeling task.
Barnes et al. \cite{DBLP:conf/acl/BarnesKOOV20} proposed to model the relationships among the subtasks by regarding this task as graph parsing.
Despite their effectiveness, these studies are insufficient to yield satisfactory performance for cross-lingual structure sentiment analysis. The key reason is the typological differences (e.g., syntactic dependencies) between the source and target languages.

Recently, multi-lingual pre-trained models such as mBERT \cite{devlin-etal-2019-bert} and XLM \cite{DBLP:conf/acl/ConneauKGCWGGOZ20} have increasingly attracted attention in
various polyglot tasks \cite{ahmad2021syntax,xu-etal-2021-xlpt}.
Mulcaire et al. \cite{mulcaire-etal-2019-polyglot} adopted multi-lingual contextualized word embedding for structure extraction tasks, such as semantic role labeling, dependency parsing, and named-entity recognition.
However, there are many versions of multi-lingual pre-trained models trained on different datasets with different training strategies so that these models contain a lot of diverse semantic and syntax information.
For example, there are more than 170 cross-lingual pre-trained models on the website\footnote{https://huggingface.co/models?sort=downloads\&search=multilingual}.
Thus, the first question (\textbf{Q1}) we want to solve is ``How to utilize these models for cross-lingual structured sentiment analysis?".

Furthermore, the existing studies have shown that the syntax structure (e.g., universal dependency tree, POS) is helpful for cross-lingual transfer \cite{subburathinam2019cross,lin-etal-2017-neural}.
Subburathinam et al. \cite{subburathinam2019cross} integrated syntax into event extraction and relation extraction and obtained a large improvement.
Also, De et al. \cite{de2021universal} released the language-universal dependency parsing across 83 languages.
This explicit syntax structure is similar among different languages since it is labeled based on the linguists' syntactic principles \cite{subburathinam2019cross}.
Also, the dependency tree structure plays a significant role in structured sentiment analysis since it can help model the relationships among the elements in the text.
Inspired by these findings, the second question (\textbf{Q2}) we want to solve is ``How to integrate the explicit structured knowledge (e.g., universal dependency tree) into cross-lingual structured sentiment analysis models?" 

To answer the above two questions, we propose a Knowledge-Enhanced Adversarial Model (\texttt{KEAM}) to integrate both the implicit distribution and explicit structure knowledge into cross-lingual structured sentiment analysis.
For \textbf{Q1}, we design an adversarial embedding adapter to obtain a robust and semantically rich word embedding from multiple pre-trained cross-lingual embedding adaptively. Particularly, a word-level attention mechanism is proposed to capture the critical semantic information from multiple embeddings, and an adversarial strategy is used to improve the robustness of the embedding.
For \textbf{Q2}, we construct a cross-lingual structure transfer learning method by integrating language-universal syntax tree via a graph convolution network (GCN) \cite{zhang-etal-2018-graph}, namely Syntax-GCN encoder. 
Finally, we obtained the structured representation used for expression extraction, target extraction, holder extraction, and sentiment classification.
In a word, to project the source language and target language into the common semantic space, we propose to share structural representations based on various pre-trained embeddings for cross-lingual structured sentiment analysis.

We conduct extensive experiments on five datasets that refer to four languages.
We compare our \texttt{KEAM} model with four supervised and four unsupervised baselines.
The proposed model outperforms all the unsupervised methods over five datasets in terms of token-level F1 and target-level F1.
The ablation studies show that both the adversarial embedding adapter and syntax-GCN encoder can improve the performance effectively.
Also, we run the transfer experiments over 25 tasks to exploit what has been learned in one language to improve generalization in another. 

The main contributions of this paper are summarized as follows.
\begin{itemize}
    \item We propose a knowledge-enhanced adversarial model for cross-lingual structured sentiment analysis by integrating both the implicit multiple embeddings and explicit syntax information.
    \item We design an adversarial embedding adapter to combine the multiple embeddings adaptively using a word-level mechanism and an adversarial strategy. Moreover, a syntax GCN is designed to integrate explicit structural knowledge.
    \item A series of experiments on five datasets that refer to four languages show the great advantage of our \texttt{KEAM} model by comparing with both supervised and unsupervised baselines in terms of various metrics.
\end{itemize}

\section{Related Work}
The key challenges for cross-lingual structured sentiment analysis tasks are complex structural sentiment extraction and cross-lingual transfer. 
Thus, we mainly review the most related works about fine-grained sentiment analysis and cross-lingual structure extraction tasks.

\subsection{Fine-grained Sentiment Analysis}
Fine-grained sentiment analysis plays an important role in sentiment analysis, including targeted sentiment analysis (TSA) \cite{mitchell2013open}, aspect-based sentiment analysis (ABSA) \cite{pontiki2016semeval,zhou2020position,zhou2019deep}, and structured sentiment analysis \cite{DBLP:conf/acl/BarnesKOOV20,wiebe2005annotating}.
TSA consists of two subtasks, target extraction, and polarity classification \cite{mitchell2013open}.
This task aims at inferring fine-grained sentiment polarity towards the targets in the given text. 
Thus, detecting the targets' sentiment based on the identified target-related contexts in the text is the biggest challenge for this task.
In recent years, neural networks have been proven effective in modeling the relationships between the target and its contexts.
To take the expression into account, ABSA is proposed to summarize users' opinions towards specific aspects in a sentence. 
ABSA not only predicts the aspect (target) and its sentiment polarities but also extracts the terms of the opinion towards the given aspect.
He et al. \cite{he2019interactive} proposed a joint approach (IMN) via a multi-task learning framework without the relationships between the target and expression.
Then, relation-aware collaborative learning (RACL) is designed to consider the relatedness \cite{chen2020relation}.

Moreover, the structured sentiment analysis task is proposed aims to extract the holders, targets, expressions, and polarities w.r.t. the given targets from the unstructured text. 
Zhang et al. \cite{zhang2019end} presented a transition-based end-to-end method to extract the elements (e.g., holders, targets, and expressions) with their relationships.
However, the sentiment information of the target is ignored by this work.
Barnes et al. \cite{DBLP:conf/acl/BarnesKOOV20} regarded the structured sentiment analysis as a dependent parsing problem to model the relatedness among the holder, target, expression, and polarity.
However, all these works focus on monolingual structured sentiment analysis.
Due to the complexity of this task, the existing datasets refer to a few languages and are relatively small, which largely limits the effectiveness of the neural network models.
In this paper, we focus on cross-lingual structured sentiment analysis tasks.

\subsection{Cross-lingual Structure Extraction}
The goal of structure extraction (e.g., information extraction, event extraction) is to extract the structural information (e.g., entities, relations, and events) from the unstructured text.
Recently, cross-lingual transfer methods are proposed for structure extraction \cite{subburathinam2019cross,lu2020cross}.
Most previous efforts working with cross-lingual structure transfer rely on parallel data \cite{qian-etal-2014-bilingual}, language-specific characteristics \cite{feng-etal-2016-language}, or machine translation \cite{zou-etal-2018-adversarial}, which requires extra resources or tools.
Feng et al. \cite{feng2018improving} and Xie et al. \cite{xie2018neural} applied cross-lingual transfer to sequence labeling task without considering the complex structures.
Wang et al. \cite{wang-etal-2018-adversarial} exploited common distribution representation space across languages for relation extraction.

Recently, multi-lingual pre-trained models (e.g., mBERT \cite{devlin-etal-2019-bert} and XLM \cite{DBLP:conf/acl/ConneauKGCWGGOZ20}) obtained great success for cross-lingual transfer \cite{mulcaire-etal-2018-polyglot}.
Liu et al. \cite{liu-etal-2019-linguistic} and Mulcaire et al. \cite{mulcaire-etal-2019-polyglot} applied multi-lingual contextualized word embedding for cross-lingual tasks, such as semantic role labeling, dependency parsing and named entity recognition. 
However, all these researches utilized one kind of word embedding one time.
In fact, there are many multi-lingual pre-trained models with various semantic information since there are trained on various datasets with various optimization objectives.
Thus, we propose an adversarial embedding adapter to combine the multiple embeddings adaptively using a word-level mechanism and an adversarial strategy.
Inspired by the existing work \cite{mulcaire-etal-2018-polyglot}, we incorporate universal dependencies \cite{prazak2017cross} for cross-lingual structured sentiment analysis. 
Particularly, the syntax parsing tree is modeled to construct multi-lingual structural representations via a GCN, which has been successfully applied to several individual monolingual NLP tasks, including relation extraction \cite{zhang-etal-2018-graph},
event detection \cite{nguyen2018graph}, ABSA \cite{zhou2020sk}.

\begin{figure}[!t]
    \centering
        \includegraphics[scale=0.34]{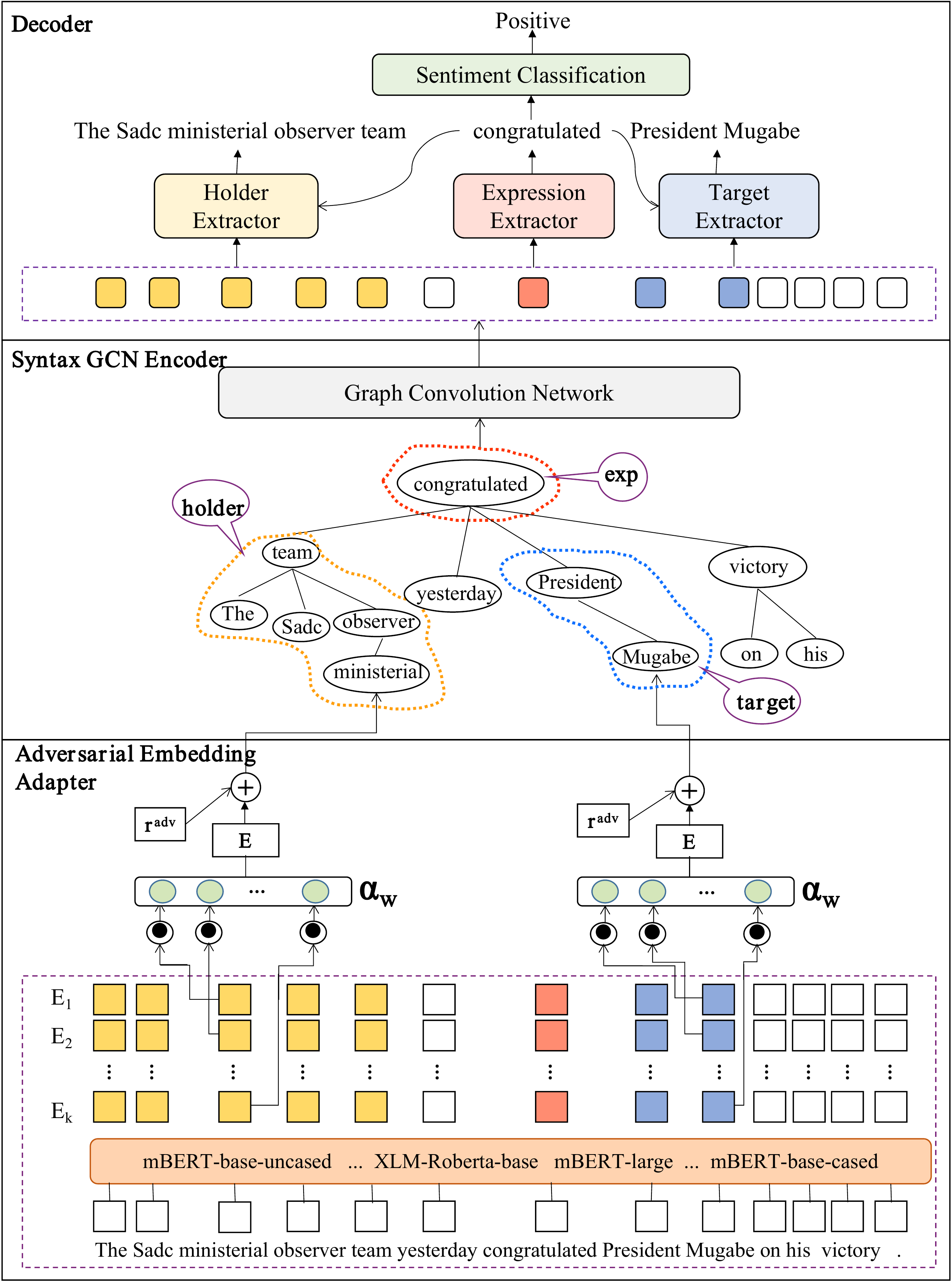}
    \caption{The framework of our \texttt{KEAM} model. It consists of three parts: adversarial embedding adapter, syntax GCN encoder, and decoder.}
    \label{fig:framework}
\end{figure}

\section{Our \texttt{KEAM} Model}
In this paper, we propose a knowledge-enhanced adversarial model for cross-lingual structured sentiment analysis by integrating both the implicit multiple embeddings and explicit syntax information. Fig. \ref{fig:framework} shows the framework of our \texttt{KEAM} model, which consists of three components, adversarial embedding adaptor, syntax GCN encoder, and decoder. First, an adversarial embedding adaptor is proposed to learn an informative and robust word embedding from multiple cross-lingual pre-trained embeddings adaptively (See Section \ref{sect:Adversarial Embedding Adapter}). Then, we design a syntax GCN encoder for learning the structural representation across different languages (See Section \ref{sect:Syntax GCN Encoder}). Finally, we perform elements extraction and sentiment classification tasks based on the sentence representation learned by the above two parts (See Section \ref{sect:decoder}).

\subsection{Task Definition}
The formulate definition of cross-lingual structured sentiment analysis is given as follows.
Given a training corpus in a source language, denoted by $\mathcal{D}_s =
\{x_1, ..., x_{|\mathcal{D}_s|}\}$, where $|\mathcal{D}_s|$ is the number of training samples in source language. 
For each sentence $x=\{w_1, ..., w_i, ..., w_{|x|}\} \in \mathcal{D}_s$, the labels $y$ is a set of opinion tuples $O = \{o_1, ..., o_{|O|}\}$.
Each opinion $o_k$ is a tuple $(h_k, t_k, e_k, p_k)$, where $h$ is a \textbf {holder} who expresses a \textbf{polarity} $p$ towards a \textbf{target} $t$ through a \textbf{sentiment expression} $e$.
Note that $h_k=\{w_{h^s_k}, ..., w_{h^e_k}\}$, $t_k=\{w_{t^s_k}, ..., w_{t^e_k}\}$, $e_k=\{w_{e^s_k}, ..., w_{e^e_k}\}$ are a sub-sequence of the sentence $x$, where ${h^s_k}$ (${h^e_k}$), ${t^s_k}$ (${t^e_k}$), ${e^s_k}$ (${e^e_k}$) are the start (end) index of the holder, target and expression, respectively. $p_k \in \{P, N\}$ denotes the positive, and negative sentiment polarities towards the given target.
Thus, cross-lingual structured sentiment analysis can be divided into four substasks: 1) expression extraction (EE) extracts the expressions $e$ in the sentence; 2) target extraction (TE) extracts the target $t$ towards the given expression $e$; 3) holder extraction (HE) extracts the holder $h$ towards the given expression $e$; 4) sentiment classification (SC) predicts the sentiment polarity $p$ towards the given expression $e$.
Finally, we aim to extract the tuple for the given sentence in the unlabelled corpus $\mathcal{D}_t$ in the target language.

\subsection{Adversarial Embedding Adapter}
\label{sect:Adversarial Embedding Adapter}
In this section, we aim to learn an informative and robust word embedding for cross-lingual transfer. First, we design a word-level attention mechanism to capture the vital distribution implicit semantic in the multiple embeddings pre-trained on a different corpus with different training strategies and tasks. Then, we adopt an adversarial training strategy to improve the robustness of the embedding.

\subsubsection{Word-level Attention}
Multi-lingual pre-trained language models, such as mBERT \cite{devlin-etal-2019-bert} and XLM \cite{DBLP:conf/acl/ConneauKGCWGGOZ20} has been widely used in different cross-lingual tasks and obtained great success.
In fact, the developer and researchers have released a large number of multi-lingual pre-trained language models. 
There are more than 100 models available on the website Hugging Face\footnote{https://huggingface.co/}.
These models have different semantic information since they are trained on different large-scale datasets using different objectives and settings.
For example, mBERT-base-cased and mBERT-base-uncased are trained on the top 104 languages with the largest Wikipedia using a masked language modeling (MLM) objective based on the lowercased or unlowercased text \cite{devlin-etal-2019-bert}.
XLM-RoBERTa model is pre-trained on the 100 languages with 2.5TB of filtered CommonCrawl data. 
Moreover, \cite{ushio-camacho-collados-2021-ner} finetuned XLM-RoBERTa models on named-entity-recognition (NER) task with multilingual datasets. 
However, most existing works use one of these models as word embedding. To obtain a better word representation, we design a word-level attention mechanism to combine the multiple embeddings adaptively.

Suppose we have $|\mathcal{M}|$ cross-lingual pre-trained models $\mathcal{M}=\{\mathcal{M}^1, \mathcal{M}^1, ..., \mathcal{M}^{|\mathcal{M}|}\}$, we input the sentence $x$ into each of them and obtained the token-level embeddings:
\begin{equation}
    E^k = \mathcal{M}^{k}(x)
\end{equation}
where $E_k=\{e^k_1, e^k_2, ..., e^k_{|x|}\}$ is the vector representations obtained by model $\mathcal{M}_k$ for sentence $x$ and $e^k_i$ means the word $w_i$'s embedding. Note that the sub-words of a word $x_i$ may be different for the different models since the tokenizers of these models are not consistent. Thus, we use the average of the sub-words embeddings in a word as the final word embedding.

Then, we design a word-level attention mechanism to combine the embedding sets $\{E^1, E^2, ..., E^{|\mathcal{M}|}\}$ into one embedding $E=\{e_1, e_2, ..., e_{|x|}\}$. We calculate the attention weight in word level because the important dimensions for different words are different. For example, the expression words focus on the sentiment information and the target words focus on the entity information. Formally, the attention weight is computed as:
\begin{equation}
\begin{aligned}
    & e_i = \sum_{k=1}^{|\mathcal{M}|}{\alpha^k_{i} \times e^k_{i}} \\
    & \alpha^k_i = \rm{softmax}(v^{T}_{a}tanh(W_{a}e^k_{i}+b_a)) \\
\end{aligned}
\end{equation}
where $v_a$, $W_a$, and $b_a$ are the trainable parameters.

\subsubsection{Adversarial Training}
The typological and semantic gaps between the source and target language lead to the instability of the cross-lingual transfer models.
Thus, to improve the robustness of the word embedding, we apply adversarial training \cite{DBLP:journals/corr/GoodfellowSS14,DBLP:conf/ijcai/SatoSS018} to input word embedding space for cross-lingual transfer.
The existing studies have shown that adversarial training is a novel regularization algorithm that improves the robustness by adding small disturbance to inputs \cite{DBLP:conf/iclr/MiyatoDG17}. 

Let $r=\{r_1, r_2, ..., r_{|x|}\}$ be a perturbation for sentence $x$, where $r_i$ is the perturbation vector for the word $w_i$.
We introduce $\hat{e}=x_{+r}=\{\hat{e}_1, \hat{e}_2, ..., \hat{e}_{|x|}\}$ that denotes $x$ with additional small perturbations $r$, where $\hat{e}_i = e_i + r_i$.
To obtain worst-case perturbations $r_{adv}$, adversarial training aims to minimize the log-likelihood,
\begin{equation}
\label{equ:r}
r^{\text{adv}}=\underset{r,\|r\| \leq \epsilon}{\operatorname{argmax}}\left\{\ell\left(x_{+r}, y\right)\right\}
\end{equation}
where $\epsilon$ is a hyper-parameter that controls the norm of the perturbation and $r^{adv}=\{r_1^{adv}, r_2^{adv}, ..., r_{|x|}^{adv}\}$.
It is generally infeasible to exactly estimate $r^{adv}$ in Equation \ref{equ:r} for sophisticated deep neural models. 
As a solution, an approximation method was proposed by linearizing $(x, y)$ around $x$ \cite{DBLP:journals/corr/GoodfellowSS14}. The approximation method induces the following
non-iterative solution for calculating $r_i^{adv}$ for all $i$.
\begin{equation}
{r}^{\text {adv}}_{i}=\frac{\epsilon {g}_{i}}{\|{g}\|_{2}}, \quad {g}_{i}=\nabla_{{e}_{i}} \ell(x, y)
\end{equation}
where $g$ is the concatenation of $g_i$ for all $i$.

Then based on adversarial perturbation $r^{adv}$, the loss function for adversarial training is computed,
\begin{equation}
\mathcal{L}_{\text{adv}}=\frac{1}{|\mathcal{D}|} \sum_{(x, y) \in \mathcal{D}} \ell\left(x_{+r^{\text {adv}}}, y\right)
\end{equation}
where $\hat{e}^{adv} = x_{+r^{adv}} = \{\hat{e}^{adv}_i, \hat{e}^{adv}_2, ..., \hat{e}^{adv}_{|x|}\}$ where $\hat{e}^{adv}_1 = e_i + r^{adv}_i$.

Finally, we jointly minimize objective functions $\mathcal{L}_{\text{task}}$
and $\mathcal{L}_{adv}$
\begin{equation}
\mathcal{L}=\mathcal{L}_{\text{task}}+\lambda \mathcal{L}_{\text {adv }}
\end{equation}
where $\mathcal{L}_{\text{task}}$ is the loss function of the structured sentiment analysis (e.g., $\mathcal{L}_{\text{exp}}$, $\mathcal{L}_{\text{target}}$, $\mathcal{L}_{\text{holder}}$ and $\mathcal{L}_{\text{polarity}}$) that will be discussed in Section \ref{sect:decoder}. $\lambda$ is used to control the balance of two terms.

\subsection{Syntax GCN Encoder}
\label{sect:Syntax GCN Encoder}
Adversarial embedding adapter focuses on learning an informative and robust distribution representation, while the explicit knowledge is ignored. 
Thus, to learn a structural representation across multiple languages, we introduce a syntax GCN encoder to integrate the dependency parsing tree into cross-lingual structured sentiment analysis. 
The main reasons are summarized as follows. 
First, the parsing tree structure plays an essential role in structured sentiment analysis. 
As shown in Fig. \ref{fig:framework}, the holder ``The Sadc ministerial observer team" and target ``President Mugabe" are both in a sub-tree. Additionally, the distance between the expression and target (or holder) on the tree is close to the sentence. 
All these show that the model can learn the structural relationships among target, holder, and expression via neighbors in the parsing trees.
Second, the tree structures about two sentences that have similar semantics are similar among multiple languages because they are labelled based on the same linguists’ syntactic principles \cite{subburathinam2019cross} while the word order is different across languages.
Thus, we propose a syntax GCN encoder to model the explicit structural knowledge for cross-lingual transfer since GCN \cite{DBLP:conf/iclr/KipfW17} is natural to model the graph.

Inspired by \cite{yao2019graph}, we construct a graph $G=(V, E)$ for each sentence $x$, where $V=\{v_1, v_2, ..., v_{|x|}\}$ is a set of nodes (words) and $E$ is a set of relations in the dependency tree. 
Here, the number of nodes in $V$ is the same as the number of words in the sentence.
Based on the graphical dependency parsing graph, we build an adjacency matrix $A \in \mathbb{R}^{|x| \times |x|}$, where $A_{ij}$ is one if there is an edge between word $w_i$ and $w_j$ in the dependency tree.
The degree matrix of $G$ is $D$ where $D_{ii}=\sum_j A_{ij}$.
Self-loops for each node are considered by adding an identity matrix to capture information about the current node itself.

The word embedding learned by the adversarial embedding adapter is input into the syntax GCN encoder.
Then, the the $l$-th graph convolution layer's hidden representation ${H}^{(l)}=\{{h}_{1}^{(l)}, {h}_{2}^{(l)}, ..., {h}_{|x|}^{(l)}\}$ is learned from its immediate neighbors' representation at the $(l-1)$-th layer. Thus, the hidden representation at the $l$-th layer is computed as:
\begin{equation}
{H}^{(l)}=\operatorname{ReLU}\left(D^{-\frac{1}{2}}{A}_{ij}D^{-\frac{1}{2}}{H}^{(l-1)}{W}^{(l)}\right)
\end{equation}
where $h^{(0)} = e$, and $W^{(l)}$, $b^{(l)}$ are the trainable parameter. Finally, we obtain the structured representations $H^{(L)}=\{{h}_{1}^{(L)}, {h}_{2}^{(L)}, ..., {h}_{|x|}^{(L)}\}$ in an universal common space from $L$-th layer, where ${h}_{i}^{(L)}$ the hidden vector of word $w_i$ in sentence $x$.

\subsection{Decoder}
\label{sect:decoder}
Based on the informative and robust structural hidden representation $H^{L}$, we adopt a simple decoder strategy for the four subtasks. First, we extract the expressions by predicting their start and end indices. 
We regard the expressions as the trigger words for each opinion. 
Then, we extract the targets and holders and predict the sentiment polarity towards the given expression. 

\paragraph{Expression Extraction}
To extract the expressions in the sentences, we predict whether the word is the start or end indices of the expressions using a two binary classifiers.
\begin{equation}
\begin{aligned}
    \boldsymbol{p}^{e^s}_i = \text{Sigmoid}(W^{e^s}h^{(L)}_i+b^{e^s}) \\
    \boldsymbol{p}^{e^e}_i= \text{Sigmoid}(W^{e^e}h^{(L)}_i+b^{e^e}) \\
\end{aligned}
\end{equation}
where $\boldsymbol{p}^{e^s}_i$ and $\boldsymbol{p}^{e^e}_i$ are the predicted expressions' start and end probabilities of the word $w_i$, $W^{e^s}$, $W^{e^e}$, $b^{e^s}$ and $b^{e^e}$ are the learnable weights. 
Cross entropy (CE) is used as the loss function,
\begin{equation}
\nonumber
\begin{aligned}
     \mathcal{L}_\text{exp}  = \frac{1}{|x|}\sum_{i=1}^{|x|}{
     {\mathrm{CE}(y^{e^s}_i, \boldsymbol{p}^{e^s}_{i})+ \mathrm{CE}(y^{e^e}_i, \boldsymbol{p}^{e^e}_{i})}
     }   
\end{aligned}
\end{equation}
where $y^{e^s}_i$/$y^{e^e}_i$  is 1 if $i$ is the start/end index of expression.

\paragraph{Target Extraction}
To take the expressions information into account for target extraction, we predict the start and end indices of the target based on the expression representations. 
\begin{equation}
\label{equ:target_p}
\begin{aligned}
    \boldsymbol{p}^{t^s_k}_i = \text{Sigmoid}(W^{t^s}[h^{(L)}_i; h^{(L)}_{e^s_k}; h^{(L)}_{e^e_k}]+b^{t^s}) \\
    \boldsymbol{p}^{t^e_k}_i = \text{Sigmoid}(W^{t^e}[h^{(L)}_i; h^{(L)}_{e^s_k}; h^{(L)}_{e^e_k}]+b^{t^e}) \\
\end{aligned}
\end{equation}
where $\boldsymbol{p}^{t^s_k}_i$ and $\boldsymbol{p}^{t^e_k}_i$ are the predicted target $t_k$'s start and end probabilities of the word $w_i$, $W^{t^s}$, $W^{t^e}$, $b^{t^s}$ and $b^{t^e}$ are the learnable weights, $h^{(L)}_{e^s_k}$ and $h^{(L)}_{e^e_k}$ are the vector representations of expression's start and end words. $[a; b]$ is the concatenation operation of $a$ and $b$.
The loss function of target extraction is,
\begin{equation}
\label{equ:target_loss}
\begin{aligned}
     \mathcal{L}_\text{target}  = \frac{1}{|O|\times|x|}\sum_{k=1}^{|O|}\sum_{i=1}^{|x|}{
     {\mathrm{CE}(y^{{t^s_k}}_i, \boldsymbol{p}^{{t^s_k}}_{i})+ \mathrm{CE}(y^{{t^e_k}}_i, \boldsymbol{p}^{{t^e_k}}_{i})}
     }   
\end{aligned}
\end{equation}
where $y^{t^s_k}_i$/$y^{t^e_k}_i$ is 1 if $i$ is the start/end index of target towards the given expression $e_k$, $|O|$ is the number of opinion tuples.

\paragraph{Holder Extraction}
The same as target extraction, we predict probabilities of the holder's start and end indices, $\boldsymbol{p}^{h^s_k}_i$ and $\boldsymbol{p}^{h^e_k}_i$, based on the expression-aware word representations (like Equation \ref{equ:target_p}). 
Then, we calculate the loss function for holder extraction, $\mathcal{L}_\text{holder}$, just like Equation \ref{equ:target_loss}.



\paragraph{Sentiment Classification}
Finally, we aim to predict the sentiment polarity $p_k$ with respect to the given expression $e_k$. We use a max-pooling operation to obtain the sentence representation $r^s=\text{Maxpooing}(H^{(L)})$ and concatenate it with expression representations for polarity classification.
\begin{equation}
\begin{aligned}
    \boldsymbol{p}^{p_k} = \text{Softmax}(W^{p}[r^s; h^{(L)}_{e^s_k}; h^{(L)}_{e^e_k}]+b^{p}) 
\end{aligned}
\end{equation}

Based on the sentiment probability distribution, we calculate the loss function as follows,
\begin{equation}
\nonumber
\begin{aligned}
    \mathcal{L}_\text{polarity}  = \frac{1}{|O|}\sum_{k=1}^{|O|}{
     {\mathrm{CE}(y^{{p_k}}, \boldsymbol{p}^{{p_k}})}
     }   
\end{aligned}
\end{equation}
where $y^{p_k}=1$ if the sentiment polarity $p_k=P$ else $y^{p_k}=0$.

\section{Experimental Setup}
\subsection{Datasets, Metrics and Implementation Details}
\paragraph{Datasets}
We evaluate our \texttt{KEAM} on five public benchmark datasets, namely NoReC$_\text{Fine}$, MultiB$_\text{EU}$, MultiB$_\text{CA}$, MPQA, DS$_\text{Unis}$, the same as \cite{DBLP:conf/acl/BarnesKOOV20}. 
NoReC$_\text{Fine}$, MultiB$_\text{EU}$ and MultiB$_\text{CA}$ are reviews in Norwegian, Basque and Catalan, respectively. 
MPQA and DS$_\text{Unis}$ are English datasets. 
We remove the samples with neutral sentiment polarities for transferring across languages since several datasets have no neutral examples.

\begin{table}[!t]
\centering
\caption{The main results of our \texttt{KEAM} model and the strong baselines. The (supervised) methods trained on the target language are marked with *. The (unsupervised) methods without * are trained on source language and tested on the target language. Note that we report the results of the source language that obtained the best transfer performance for these unsupervised methods. The best results of supervised and unsupervised methods are marked with \underline{underline} and \textbf{bold}.}
\label{table:main results}
\setlength{\tabcolsep}{0.4mm}{\begin{tabular}{ll|cccc}
\hlineB{4}
\multirow{2}{*}{Target}          & \multicolumn{1}{c|}{\multirow{2}{*}{Methods}} & \multicolumn{3}{c}{Token}                                                                    & \multicolumn{1}{c}{Targeted F1} \\
                                 & \multicolumn{1}{c|}{}                         & \multicolumn{1}{c}{Holder F1} & \multicolumn{1}{c}{Target F1} & \multicolumn{1}{c}{Exp. F1} & \multicolumn{1}{c}{F1}          \\
                                 \hline
\multirow{12}{*}{NoReC$_\text{Fine}$} & Ovrelid et al. \cite{ovrelid-etal-2020-fine}*                                        & 42.4                              &    31.3                           &     31.1                        &               -                  \\
                                 & IMN*                                         &  -                              &  35.9                             &    48.7                          &  18.0                               \\
                                 & RACL-BERT*                                   &  -                             &  47.2                        &   56.3                         &   30.3                               \\
                                 & RACL*                                        & -                              &  45.6                           & 55.4                           &    20.1                              \\
                                 & Head-first*                                  &  51.1                             & \underline{50.1}                              &                       \underline{54.4}      &   \underline{30.5}                              \\
                                 \cline{2-6} 
                                 & mBERT$_\text{base}$                                   &                       26.7   & 17.4   & 26.0 & 10.9  \\
                                 & XLM-RoBERTa$_\text{base}$                             &                         0.0    & 20.5   & 26.7 & 14.0  \\
                                 & XLM-RoBERTa$_\text{large}$                            &                     2.9    & 20.1   & 27.9 & 15.6 \\
                                 & XLM-RoBERTa$_\text{large}^{F}$                      &            34.5   & 31.6   & 28.9 & 22.2      \\ \cline{2-6} 
                                 & \texttt{KEAM}*  & \underline{55.7}   & 41.8   & 30.0 & 29.1    \\
                                 & \texttt{KEAM}                                 &     \textbf{41.9}   & \textbf{39.8}   & \textbf{30.7} & \textbf{25.2}    \\ 
                                 \hline
\multirow{12}{*}{MultiB$_\text{CA}$} & Barnes et al. \cite{barnes2018multibooked}*                                          &   56.0                            &    64.0                           &  52.0                           &   -                              \\
                                 & IMN*                                         &     -                          &   56.3                            &   60.9                         &   32.5                              \\
                                 & RACL-BERT*                                   &     -                          &  67.5                             &    70.3                         &   52.4                              \\
                                 & RACL*                                        &  -                             &  65.4                            & 67.6                             &   49.1                              \\
                                 & Head-first*                                  &   43.0                          & \underline{72.5}                             &      \underline{71.1}                       & \underline{55.0}                              \\
                                 \cline{2-6} 
                                & mBERT$_\text{base}$                                   &                  0.0    & 5.9    & 51.5 & 6.2                          \\
                                 & XLM-RoBERTa$_\text{base}$                             &                0.0    & 25.3   & 53.4 & 17.8   \\
                                 & XLM-RoBERTa$_\text{large}$                            &             13.6   & 40.7   & 52.0 & 22.5   \\
                                 & XLM-RoBERTa$_\text{large}^{F}$                      &                         9.5    & 41.6   & 51.7 & 23.2  \\ \cline{2-6} 
                                 & \texttt{KEAM}*                                  &         \underline{64.0}   & 67.8   & 58.5 & 38.7     \\
                                 & \texttt{KEAM}                                  & \textbf{32.7}   & \textbf{48.1}   & \textbf{54.9} & \textbf{26.7}       \\ 
                                 \hline
\multirow{12}{*}{MultiB$_\text{EU}$} & Barnes et al. \cite{barnes2018multibooked}*                                        &   54.0                            &   57.0                            &   54.0                          &    -                             \\
                                 & IMN*                                         &   -                            &    48.2                           &     65.2                        &   39.5                              \\
                                 & RACL-BERT*                                   & -                              &  59.9                             &           72.6                  &       56.8                          \\
                                 & RACL*                                        & -                              &  55.4                             &     70.7                        &   48.2                              \\
                                 & Head-first*                                  &  60.4                             & 64.0                              &            \underline{73.9}                 &      \underline{57.8}                           \\
                                \cline{2-6} 
                                 & mBERT$_\text{base}$                                   &                 0.0    & 31.9   & 46.7 & 30.6  \\
                                 & XLM-RoBERTa$_\text{base}$                             &                 3.4    & 50.3   & 46.9 & 43.7  \\
                                 & XLM-RoBERTa$_\text{large}$                            &                   0.0    & 51.8   & 45.5 & 43.4 \\
                                 & XLM-RoBERTa$_\text{large}^{F}$                      &  0.0    & 52.3   & 47.6 & 47.0 \\ \cline{2-6} 
                                 & \texttt{KEAM}*                                  &              \underline{64.6}   & \underline{65.3}   & 62.2 & 52.7     \\
                                 & \texttt{KEAM}              & \textbf{3.8}    & \textbf{58.6}   & \textbf{50.4} & \textbf{48.0}      \\ 
                                 \hline
\multirow{11}{*}{MPQA} 
                                 & IMN*                                         &  -                             &24.3                               & 29.6                            &    1.2                             \\
                                 & RACL-BERT*                                   &   -                            & 20.0                               &   31.2                          &  17.8                               \\
                                 & RACL*                                        &  -                             &32.6                               &   37.8                          &  11.8                               \\
                                 & Head-first*                                  &       \underline{43.8}                        &  \underline{51.0}                             &     \underline{48.1}                        &    \underline{33.5}                             \\
                                 \cline{2-6} 
                                 & mBERT$_\text{base}$                                   &       0.4                        &   2.3                            &     9.7                        &           1.1                      \\
                                 & XLM-RoBERTa$_\text{base}$                             &      4.6                         &       3.4                        &       9.6                      &         2.1                        \\
                                 & XLM-RoBERTa$_\text{large}$                            &   4.5                            &          4.8                     &       10.6                      &          1.9                       \\
                                 & XLM-RoBERTa$_\text{large}^{F}$                      &      10.5                         &        5.0                       &     40.7                        &      2.2                           \\ \cline{2-6} 
                                 & \texttt{KEAM}*                                  & 34.2   & 45.6   & 37.8 & 15.2   \\
                                 & \texttt{KEAM}                                  &  \textbf{12.3}   & \textbf{5.7}    & \textbf{12.8} &\textbf{ 2.7 }  \\ 
                                 \hline
\multirow{11}{*}{DS$_\text{Unis}$} 
                                 & IMN*                                         &      -                         &   33.0                            &    27.4                         &   17.9                              \\
                                 & RACL-BERT*                                   &    -                           &  44.6                             &       38.2                      &  \underline{27.3}                               \\
                                 & RACL*                                        &      -                         &  39.3                            &     40.2                       & 22.8                                \\
                                 & Head-first*                                  &      28.0                         &  39.9                             &  \underline{40.3}                           &  26.7                               \\
                                 \cline{2-6} 
                                 & mBERT$_\text{base}$                                   &      1.6             & 26.1   & 16.2 & 8.0  \\
                                 & XLM-RoBERTa$_\text{base}$                             &      3.0                        & 23.6   & 17.7 & 6.3 \\
                                 & XLM-RoBERTa$_\text{large}$                            &    3.4                        & 27.8   & 17.6 & 10.4                 \\
                                 & XLM-RoBERTa$_\text{large}^{F}$                      &    4.5               & 28.8   & 17.4 & 9.4                   \\ \cline{2-6} 
                                 & \texttt{KEAM}*                                 & \underline{31.2}   & \underline{44.7}   & 36.4 & 21.5   \\
                                 & \texttt{KEAM}                        & \textbf{8.1}  & \textbf{30.4}   & \textbf{18.5} & \textbf{12.4}      \\      
\hlineB{4}
\end{tabular}}
\end{table}

\begin{table}[!t]
\centering
\caption{The results of ablation studies. AEA: Adversarial Embedding Adapter. SGCNE: Syntax-GCN Encoder. The average score on the target language are reported.}
\label{table:ablation study}
\setlength{\tabcolsep}{0.6mm}{\begin{tabular}{ll|cccc}
\hlineB{4}
\multirow{2}{*}{Target}          & \multicolumn{1}{c|}{\multirow{2}{*}{Methods}} & \multicolumn{3}{c}{Token}                                                                    & \multicolumn{1}{c}{Targeted F1} \\
                                 & \multicolumn{1}{c|}{}                         & \multicolumn{1}{c}{Holder F1} & \multicolumn{1}{c}{Target F1} & \multicolumn{1}{c}{Exp. F1} & \multicolumn{1}{c}{F1}          \\
                                 \hline
\multirow{3}{*}{NoReC$_\text{Fine}$}  & \texttt{KEAM}     &   \textbf{16.6} & \textbf{32.3} & \textbf{28.0 }& \textbf{16.8}    \\
 & - AEA     &     14.3 & 29.2 & 26.2 & 16.3\\
  & - SGCNE       &     13.4 & 30.3 & 27.1 & 14.1  \\
                                 \hline
\multirow{3}{*}{MultiB$_\text{CA}$}  & \texttt{KEAM}                                 &         \textbf{12.9} & \textbf{43.0} & \textbf{49.4} & \textbf{14.3} \\
 & - AEA          &    11.5 & 37.8 & 48.3 & 11.7 \\
  & -  SGCNE         &  10.7   &  42.1 &  47.1 & 13.9 \\
                                 \hline
\multirow{3}{*}{MultiB$_\text{EU}$}  & \texttt{KEAM}                                 & \textbf{5.2} & \textbf{46.8} & \textbf{45.8} & \textbf{17.0} \\
 & - AEA        &  3.9 & 39.0 & 45.6 & 14.9 \\
  & - SGCNE        &   4.4           &     42.5                    &    41.9             &  15.5 \\
                                 \hline
\multirow{3}{*}{MPQA}  & \texttt{KEAM}                                 &                          \textbf{10.4} & \textbf{7.6} & \textbf{12.1} & \textbf{2.3}                     \\
 & - AEA   &   8.7 & 7.2 & 12.0 & 1.7\\
  & - SGCNE       &     9.1                    &   7.3                       &     11.5              &  2.0 \\
                                 \hline
\multirow{3}{*}{DS$_\text{Unis}$}  & \texttt{KEAM}                                 &      \textbf{2.0} & \textbf{26.97} & \textbf{18.5} & \textbf{11.8} \\
 & - AEA                                &     0.8                          &      20.0                         &       17.1                      &   9.8  \\
  & - SGCNE                                 &   1.6                            &      23.6                         &    18.1                    &   10.5    \\
\hlineB{4}
\end{tabular}}
\end{table}

\paragraph{Metrics}
Following the previous works \cite{DBLP:conf/acl/BarnesKOOV20}, we adopt four metrics from token and target levels. The token-level metrics are holder F1 (Holder F1), target F1 (Target), and expression F1 (Exp. F1).
For target-level metrics, we utilize targeted F1, which is commonly used in targeted sentiment analysis \cite{he2019interactive}. Notably, a correct prediction requires that both the target boundary and its sentiment polarity be correct.

\paragraph{Implementation Details}
Adam optimizer is utilized with the learning rates of 1e-5. The max sequence length is set as 128. The dropout is 0.1. We implement our \texttt{KEAP} model based on the transformers framework \footnote{https://huggingface.co/transformers/}. 
We use Stanza \footnote{https://github.com/stanfordnlp/stanza} tool to obtain the universal dependencies tree.
The default number of GCN layers is 3 in our experiments.
We train our proposed model for 50 epochs, run it five times with
different random seeds and report the mean.
The reported test results are based on the model that performs best regarding targeted F1 on the development set.

\subsection{Baselines}
To evaluate the effectiveness of our \texttt{KEAM} model, we compare our approach with the state-of-the-art baselines from two views, supervised and unsupervised (transfer) methods.

The following typical supervised baselines are selected,
1) IMN \cite{he-etal-2019-interactive} performs token-level classification with document-level classification.
2) RACL \cite{chen2020relation} utilizes a is a relation propagation mechanism via a multi-task learning strategy for the subtasks of aspect-based sentiment analysis.  
3) RACL-BERT \cite{chen2020relation} is an extension of RACL that replaces embedding with BERT-Large.
4) Head-first \cite{DBLP:conf/acl/BarnesKOOV20} translates the structured sentiment analysis into a dependency parsing problem and obtains the state-of-the-art results.
We also compare with previously reported extraction results from Barnes et al. \cite{barnes2018multibooked} and Ovrelid.
et al. \cite{ovrelid-etal-2020-fine}.

For the unsupervised (transfer) methods, four typical multi-lingual pre-trained language models  \cite{devlin-etal-2019-bert,DBLP:conf/acl/ConneauKGCWGGOZ20} are utilized as the baselines.
These models have different semantic information since they are trained on different large-scale datasets using different objectives and settings.
First, mBERT$_\text{base}$ is trained on the top 104 languages with the largest Wikipedia using a masked language modeling (MLM) objective \cite{devlin-etal-2019-bert}.
XLM-RoBERTa$_\text{base}$ and XLM-RoBERTa$_\text{large}$ models are pre-trained on the 100 languages with 2.5TB of filtered CommonCrawl data. 
Moreover, Ushio et al. \cite{ushio-camacho-collados-2021-ner} finetuned XLM-RoBERTa$_\text{large}$ model on named-entity-recognition (NER) task with multilingual datasets, namely XLM-RoBERTa$_\text{large}^F$. 

\section{Experimental Results}
\subsection{Main Results}
To verify the effectiveness of our \texttt{KEAM} model, we evaluate it on a cross-lingual transfer setting that trains on source language and tests on the target language without labelled data. 
We compare our model with supervised and unsupervised baselines (See Table \ref{table:main results}).
Note that we report the results on the source dataset that obtains the best performance on the target language for the unsupervised transfer methods (without *).

From this table, we obtain the following observations. 
\textbf{First}, our transfer model (\texttt{KEAM}) performs better than the unsupervised baselines significantly for cross-lingual structured sentiment analysis.
To be specific, our model obtains the best performance overall the datasets in terms of all the metrics. Moreover, \texttt{KEAM} improves more than three points on three datasets in terms of targeted F1.
All these indicate that our informative structural representation can help model transfer structural sentiment across languages.
Moreover, the performance of various multi-lingual embeddings is very different due to various multi-lingual embeddings having different semantic information.
\textbf{Second}, the performance of \texttt{KEAM} is comparable with the supervised methods (e.g., \texttt{KEAM}*, IMN) in some cases. 
For example, the proposed model performs better than IMN over NOReC$_\text{Fine}$ and MultiB$_\text{CA}$ regarding targeted F1.
\textbf{Third}, our model also performs well on the supervised tasks (\texttt{KEAM}*). We train our model on the target language with a supervised method. We observe that \texttt{KEAM}* performs better or is comparable with the supervised baselines in most cases.
\textbf{Fourth}, regrading holder F1, \texttt{KEAM} (\texttt{KEAM}*) outperforms better than the unsupervised (supervised) methods in most cases.
The frequency of holders is relatively low, which makes it hard to extract the holders in the text.
We incorporate syntax information via a GCN model to a structural representation, which helps \texttt{KEAM} identify the holders.

\subsection{Ablation Studies}
Furthermore, to investigate the effectiveness of each component consisting of our \texttt{KEAM}, we do the ablation test over five datasets (See Table \ref{table:ablation study}). 
The average score over all the source datasets except the target dataset is reported.
Specifically, we remove the adversarial embedding adapter (-AEA) and syntax-GCN encoder (-SGCNE) from the \texttt{KEAM} respectively.
For -AEA, we simply use the concatenation of multiple multi-lingual word embeddings without adversarial training.

The results demonstrate that both adversarial embedding adapter and syntax-GCN encoder are important for this task.
First, our adversarial embedding adapter can capture the diversity of essential features from various multi-lingual embeddings that contain different semantic information. 
It learns an informative embedding via an attention mechanism and improves the robustness via an adversarial training strategy.
The final word embeddings can improve the performance of cross-lingual transfer well. 
Second, the structural representations learned by the syntax GCN encoder can further improve the model effect. It is because the dependency tree is vital for structured sentiment analysis, and the tree structures across languages are similar.

\subsection{The Effectiveness of Transfer Learning}
In this section, we basically try to exploit what has been learned in one language to improve generalization in another.
Table \ref{table:transfer} shows the results of transfer learning across languages in terms of targeted F1.
The findings are summarized as follows. 
1) The source language will influence the performance largely.
For example, transferring knowledge from MultiB$_\text{CA}$ obtains outstanding performance for MultiB$_\text{EU}$ while having a limited effect for MPQA.
2) Transferring from the target language may not obtain the best scores. 
For instance, though both MPQA and DS$_\text{Unis}$ are English datasets, transferring from NoReC$_\text{Fine}$ obtains the best performance for them.
3) It is obvious that training on the target dataset (supervised approaches) obtains the best results since the train and test datasets are in the same distribution.

\begin{table}[!t]
\centering
\caption{The performance of transfer learning in terms of targeted F1.}
\label{table:transfer}
\setlength{\tabcolsep}{0.8mm}{\begin{tabular}{l|ccccc}
\hlineB{4}
               & \multicolumn{1}{c}{\textbf{NoReC$_\text{Fine}$}} & \multicolumn{1}{c}{\textbf{MultiB$_\text{CA}$}} & \multicolumn{1}{c}{\textbf{MultiB$_\text{EU}$}} & \multicolumn{1}{c}{\textbf{MPQA}} & \multicolumn{1}{c}{\textbf{DS$_\text{Unis}$}} \\ \hline
\textbf{NoReC$_\text{Fine}$} &       \underline{29.1}                   &                               6.5             &       7.7                                     &               \textbf{3.7}                   &                           \textbf{16.8}               \\
\textbf{MultiB$_\text{CA}$}  &           \textbf{25.2}                                   &         \underline{38.7}                                   &                      \textbf{48.0}                      &      1.7                            &                12.4                          \\
\textbf{MultiB$_\text{EU}$}  &         24.5                                    &           \textbf{26.7}                                 &                        \underline{52.7}                    &        1.1                           &            10.9                              \\
\textbf{MPQA}           &      6.5                                       &                 11.6                           &                            5.0                &         \underline{15.2}                          &               8.9                           \\
\textbf{DS$_\text{Unis}$}    &      11.2                                       &           5.9                                 &                       7.2                     &   2.7                                &  \underline{21.5} \\
\hlineB{4}                                       
\end{tabular}}
\end{table}

\section{Conclusions and Future Work}
In this paper, we propose \texttt{KEAM} for cross-lingual structured sentiment analysis. 
Mainly, we first design an adversarial embedding adaptor for learning an informative and robust embedding. Then, a syntax GCN encoder is introduced to learn structural representation based on the parsing tree. We compare our model with both supervised and unsupervised baselines on five datasets with four languages. The experimental results show the great advantage of our model for cross-lingual transfer. We also do the ablation studies to investigate the effectiveness of each part in our model. 
To investigate the effectiveness of cross-lingual transfer, we run 25 transfer tasks across five datasets.

In future work, we would like to explore more multi-lingual pre-trained models. Also, exploring the performance of our model for other structure extraction tasks is interesting.

\section*{Acknowledgment}
The authors wish to thank the reviewers for
their helpful comments and suggestions. 
This research is funded by the Science and Technology Commission of Shanghai Municipality (No. 19511120200 and 21511100402) and by Shanghai Key Laboratory of Multidimensional Information Processing, East China Normal University, No. 2020KEY001.
This research is also funded by the National Key Research and Development Program of China (No. 2021ZD0114002), the National Nature Science Foundation of China (No. 61906045), and Shanghai Science and Technology Innovation Action Plan International Cooperation project ``Research on international multi language online learning platform and key technologies (No.20510780100)". The computation is performed in ECNU Multi-functional Platform for Innovation (001).

\small
\bibliographystyle{IEEEtran}
\bibliography{main}

\end{document}